\documentclass[conference]{IEEEtran}
\IEEEoverridecommandlockouts
\usepackage{hyperref}
\usepackage{cite}
\usepackage{amsmath,amssymb,amsfonts}
\usepackage{algorithmic}
\usepackage{graphicx}
\usepackage{textcomp}
\usepackage{xcolor}
\usepackage{multirow} 
\def\BibTeX{{\rm B\kern-.05em{\sc i\kern-.025em b}\kern-.08em
    T\kern-.1667em\lower.7ex\hbox{E}\kern-.125emX}}
\begin{document}

\title{Ideology as a Problem: Lightweight Logit Steering for Annotator-Specific Alignment in Social Media Analysis}

\author{

    \IEEEauthorblockN{
        Wei Xia\IEEEauthorrefmark{1}\IEEEauthorrefmark{2}, 
        Haowen Tang\IEEEauthorrefmark{1}\IEEEauthorrefmark{2}, 
        and Luozheng Li\IEEEauthorrefmark{1}\IEEEauthorrefmark{3}
    }

    \IEEEauthorblockA{\IEEEauthorrefmark{1}Institute of Software, Chinese Academy of Sciences (ISCAS)}
    \IEEEauthorblockA{\IEEEauthorrefmark{2}Ludwig Maximilian University of Munich}

    \IEEEauthorblockA{
        Emails: W.Xia1@campus.lmu.de, Haowen.Tang@campus.lmu.de, liluozheng@iscas.ac.cn
    }

    \thanks{\IEEEauthorrefmark{3}Corresponding author: Luozheng Li (Assistant Researcher).}
}

\maketitle

\begin{abstract}
LLMs internally organize political ideology along low-dimensional structures that are partially—but not fully—aligned with human ideological space.
This misalignment is systematic, model-specific, and measurable.
We introduce a lightweight linear probe that both quantifies the misalignment and minimally corrects the output layer.This paper introduces a simple and efficient method for aligning models with specific user opinions. Instead of retraining the model, we calculated a bias score from its internal features and directly adjusted the final output probabilities. This solution is practical and low-cost and preserves the original reasoning power of the model.
\end{abstract}


\section{Introduction}
\label{sec:intro}
Large Language Models (LLMs) are increasingly applied to political analysis, content moderation, and computational social science. In these settings, ideological neutrality is essential. However, recent studies show that LLMs often exhibit systematic political preferences, typically leaning left-liberal across a wide range of topics \cite{santurkar2023whose, durmus2023towards, rozado2024political}. These tendencies are partly inherited from the distributions of online data on which the models are trained. As a result, the political judgments produced by LLMs may diverge from those of individual human annotators, especially when annotators hold different ideological positions.

This mismatch creates a practical challenge. Political annotation is inherently subjective: different annotators may interpret the same text from different ideological standpoints. When an LLM is used to assist or replace human annotators, a fixed ideological bias in the model can systematically under-represent certain viewpoints and reduce annotation quality.

Existing mitigation strategies are limited. Full-model fine-tuning (e.g., RLHF) \cite{ouyang2022traininglanguagemodelsfollow} requires substantial data and computational resources and is impractical for adapting to many different annotators. Representation-level interventions attempt to steer internal activations \cite{zou2025representationengineeringtopdownapproach}, but such methods may introduce unintended side effects and degrade the model's general capabilities \cite{turner2024steeringlanguagemodelsactivation}. This motivates the search for lightweight approaches that can adjust ideological outputs without modifying the underlying model.

In this work, we take a different perspective. Rather than treating ideological bias as a monolithic property of the entire model, we examine how ideological information is encoded inside hidden representations and how it is read out by the final layer. Our analyses suggest that the model’s internal representations contain meaningful ideological structure \cite{feng2023pretraining}, but the mapping from these representations to discrete left/center/right predictions is not always aligned with human judgment. This observation motivates a simple readout-level correction mechanism that adjusts the final logits without altering the model parameters.

We evaluate this approach on the MITweet dataset \cite{liu-etal-2023-ideology}, covering twelve political facets. The method consistently improves predictive accuracy and reduces systematic misalignment across multiple LLMs, including Llama3 and Qwen.

Our contributions are as follows:
\begin{itemize}
    \item We present an empirical analysis of how LLMs encode political ideology across multiple facets, revealing a consistent low-dimensional structure in hidden representations.
    \item Based on this observation, we introduce a lightweight, non-invasive logit-level adjustment mechanism that aligns model outputs with annotator-specific ideological perspectives.
    \item We demonstrate the effectiveness of this approach across twelve political facets and multiple LLMs, and release our implementation to facilitate further research on personalized and fair political annotation.
\end{itemize}

\section{RELATED WORK}

The study of bias in NLP has progressed from static word embeddings \cite{bolukbasi2016man} to modern LLMs. A growing body of work shows that LLMs do not represent a neutral “view from nowhere,” but instead tend to reflect liberal or Western-centric perspectives present in their pretraining data \cite{santurkar2023whose, feng2023pretraining}. This creates tension when models are used to assist human annotators with diverse ideological backgrounds. Political science literature has long emphasized that ideology is multi-dimensional 
rather than a single left–right continuum \cite{feldman2014understanding, kennedy2020constructingintervalvariablesfaceted}.
The “Value Kaleidoscope’’ framework \cite{sorensen2024value} emphasizes that alignment is inherently pluralistic: human annotators form dissenting ideological clusters \cite{davani-etal-2022-dealing}, and models optimized for average satisfaction can miss the preferences of specific groups \cite{gordon2022jury}. These findings suggest that a uniform, one-size-fits-all LLM is poorly suited for tasks requiring granular simulation of ideological viewpoints.

Existing approaches to mitigating ideological misalignment fall roughly into two categories. Training-based alignment methods—such as RLHF and its variants \cite{ouyang2022traininglanguagemodelsfollow, rafailov2024directpreferenceoptimizationlanguage}—require substantial data and computation. Although work on diverse preference modeling \cite{bakker2022fine} seeks to capture broader viewpoints, such approaches remain expensive and are not designed for rapid adaptation to individual annotators. A second category centers on prompting and auditing. Studies have quantified LLM political leanings \cite{rozado2023political, hartmann2023politicalideologyconversationalai}, and persona-based prompting \cite{Argyle_Busby_Fulda_Gubler_Rytting_Wingate_2023} can shift outputs, but results are often brittle and rely on superficial stereotypes rather than genuine ideological grounding \cite{tjuatja2024llmsexhibithumanlikeresponse}. These limitations make such methods unreliable for fine-grained ideological simulation.

A complementary line of work examines the geometry of LLM representations. Sociological analyses further show that political ideology can emerge as an ordered 
geometric axis in distributional embedding spaces \cite{Kozlowski_2019}, 
suggesting that ideological structure may be captured by low-dimensional directions.
Several studies suggest that certain high-level concepts—such as sentiment or truthfulness—may be encoded along approximately linear directions in the hidden space \cite{mamou2020emergenceseparablemanifoldsdeep, park2023linear, burns2022discovering}.Such structure is consistent with prior findings that transformer representations 
are highly anisotropic, with major semantic variation concentrated along a few 
dominant directions \cite{timkey2021barkbiteroguedimensions}.
 Building on this perspective, Zou et al.\ propose Representation Engineering (RepE) \cite{zou2025representationengineeringtopdownapproach}, which modifies activations by injecting steering vectors. While effective, these interventions are invasive and may introduce unintended side effects on general capabilities \cite{turner2024steeringlanguagemodelsactivation}. Logit-based approaches provide a non-invasive alternative: for example, DoLa \cite{chuang2024doladecodingcontrastinglayers} adjusts logits by contrasting layers to improve factuality. However, existing logit-level methods focus primarily on hallucination or broad safety criteria. To our knowledge, the combination of geometric insights from representation space with lightweight, logit-level calibration has not yet been systematically explored for pluralistic ideological alignment.

\section{Methodology}
\label{sec:method}

\begin{figure*}[t]
    \centering
    \includegraphics[width=\textwidth]{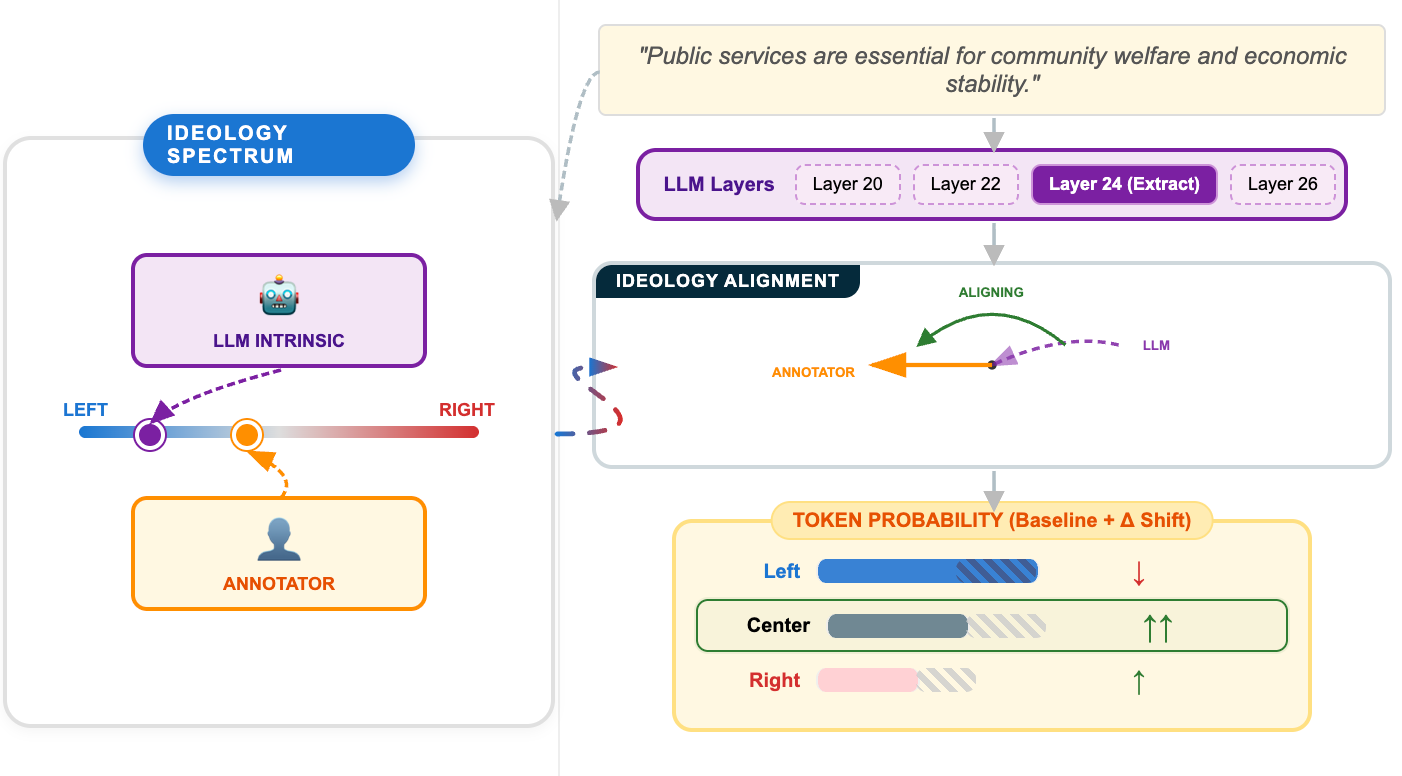}
    \caption{\textbf{Overview of the Proposed Logit-Steering Framework.}
    Given an input text, the frozen LLM produces a hidden representation $h$.
     And the asymmetric update rule modulates the original logits to
    align predictions with annotator-specific ideological preferences. Only a
    handful of scalar parameters are trained, while the LLM remains frozen.}
    \label{fig:pipeline}
\end{figure*}

Our goal is to adapt a frozen LLM so that its ideological predictions
(\textit{Left / Center / Right}) align with the preferences of a target annotator.
The method operates entirely at the logit level and introduces only a small number
of trainable scalar parameters, keeping the underlying LLM unchanged. Figure~\ref{fig:pipeline} illustrates the full workflow of our method.

\subsection{Problem Setup}

Given an input text $x$, a frozen LLM produces a hidden representation
$h \in \mathbb{R}^d$ from a chosen transformer layer and an unnormalized logit vector
\[
z = [z_L, z_C, z_R] \in \mathbb{R}^3.
\]
We assume access to a small annotator-specific dataset
\[
\mathcal{D}_{\mathrm{few}} = \{(x_i, y_i)\},
\]
where $y_i \in \{\text{Left},\text{Center},\text{Right}\}$.
The objective is to learn a minimal correction mechanism mapping $z$ to calibrated logits $\hat{z}$.

\subsection{Hidden State Extraction}

For each input $x$, we extract the hidden representation $h$ from a fixed transformer layer
(e.g., the final layer).  
The LLM parameters remain frozen throughout training.  
These hidden states provide a stable semantic basis for downstream calibration.

\subsection{Dual-Probe Decomposition}
To capture the distinct geometric properties of steering, we decompose the correction mechanism into two scalar components: a directional term $s$ and a score $g$.

\begin{equation}
s = \mathbf{v}_s^\top \mathbf{h} + b_s
\end{equation}

\begin{equation}
g = \text{Softplus}(\mathbf{v}_g^\top \mathbf{h} + b_g)
\end{equation}

\paragraph{Directional term ($s$).}
The term $s$ captures the signed Left--Right tendency encoded in $h$.  
It determines in which direction the logits should be shifted when redistribution is needed.

\paragraph{Score ($g$).}
The term $g$ provides a non-negative measure of how strongly the logits should be adjusted.  
Using the Softplus activation ensures $g \ge 0$, so that $g$ contributes only a magnitude and does not introduce an additional direction.  
This separates directional information (handled by $s$) from the size of the correction (handled by $g$), allowing the model to adjust $z_C$ when the representation points toward ideological ambiguity.

\subsection{Asymmetric Logit Calibration}
\label{sec:calib}

We combine the directional term $s$ and the score $g$ through an asymmetric update of the logits:
\begin{align}
\hat{z}_L &= z_L - s - \tfrac12 g, \\
\hat{z}_C &= z_C + \mu g, \\
\hat{z}_R &= z_R + \mu s - \tfrac12 g.
\end{align}

This formulation separates three roles: direction, magnitude, and redistribution.

\paragraph{Symmetric reduction of polarized logits.}
The score $g$ represents the amount of correction applied to the polarized classes.
Subtracting $\tfrac12 g$ from both $z_L$ and $z_R$ reduces their influence in a balanced way, independent of the sign of $s$.
This makes it possible for the Center class to become competitive in cases where the original logits favor polarized outcomes too strongly.

\paragraph{Redistribution controlled by $\mu$.}
The coefficient $\mu \in [0,1]$ determines how much of the reduced amount is added back to other classes.
A fraction $\mu g$ is assigned to the Center logit, and a fraction $\mu s$ adjusts the Left--Right balance.
When $\mu < 1$, the update remains conservative: the model does not assume that all removed mass should be reassigned, which prevents overly large shifts in ambiguous inputs.

\paragraph{Overall effect.}
The term $s$ determines the direction of adjustment, $g$ determines its scale, and $\mu$ controls the strength of redistribution.
Together, they allow the calibration to both counteract overconfident polarized predictions and promote the Center class when the representation indicates uncertainty or neutrality.

\subsection{Training Objective}

Only the parameters
\[
\theta = \{v_s, b_s, v_g, b_g, \mu\}
\]
are trained, while the LLM remains frozen.
We optimize the cross-entropy loss:
\[
\mathcal{L}(\theta)
=
- \sum_{(x,y) \in \mathcal{D}_{\mathrm{few}}}
\log p_\theta(y \mid x),
\]
where $p_\theta$ is obtained by applying a softmax to the calibrated logits $\hat{z}$.

The parameter count is extremely small, enabling fast and stable learning from
a few labeled examples while preserving the base model’s language modeling behavior.

\section{Experiments}
\label{sec:experiments}

This section evaluates our method on the MITweet benchmark, comparing it against strong prompting baselines and supervised fine-tuned PLMs. We further provide per-facet analysis, interpretive diagnostics, and a safety comparison against internal activation steering.

\subsection{Experimental Setup}
\label{subsec:setup}

\paragraph{Dataset.}
We use the MITweet dataset~\cite{liu-etal-2023-ideology}, a multifaceted benchmark for ideological stance detection. Each tweet is annotated along twelve facets (e.g., \textit{Migration}, \textit{Diplomatic Strategy}, \textit{State Structure}), with labels drawn from \{\textit{Left, Center, Right}\}. This facet structure makes the task more challenging than conventional binary stance detection, since ideological position varies systematically across domains.

\paragraph{Models.}
We evaluate two open-weights LLMs: \textbf{Qwen-2.5-7B}~\cite{yang2025qwen3} and \textbf{Llama-3-8B}~\cite{dubey2024llama}. Both models remain completely frozen; only our lightweight steering parameters are trained.

\paragraph{Metrics.}
Following prior work, we report \textbf{Accuracy} and \textbf{Macro-F1}.  
Macro-F1 is particularly informative due to class imbalance and because it penalizes the “majority-class collapse’’ (e.g., always predicting \textit{Left}) frequently observed in political tasks.

\subsection{Baselines}
\label{subsec:baselines}

We compare against two families of baselines: fully fine-tuned PLMs and prompting-based LLM inference.

\paragraph{1) Supervised PLMs.}
We include results reported in Liu et al.~\cite{liu-etal-2023-ideology} for BERT-base and BERTweet.  
These models offer a meaningful reference point for supervised learning under a non-frozen setting.

\paragraph{2) LLM Prompting.}
Using the same Qwen/Llama backbones, we evaluate:
\begin{itemize}
    \item \textbf{Zero-shot}: Direct instruction prompting without examples.
    \item \textbf{Few-shot ICL}: Five in-context demonstrations per class.
    \item \textbf{Schema-Aware Prompting}: Descriptions of the facet-specific meanings of Left/Center/Right are injected into the prompt.
\end{itemize}

\paragraph{3) Ours (Logit Steering).}
Our method learns a single vector $v$ and bias $b$ using only 20\% MITweet labeled examples per facet.  
The underlying LLM is kept frozen.

\subsection{Main Results}

Table~\ref{tab:main_results} reports performance on the full MITweet benchmark.
Two patterns emerge:
\textbf{(1) Prompting shows an Accuracy–F1 trade-off.}
Schema-aware prompting improves Accuracy but often reduces Macro-F1, indicating that the model continues to rely on its dominant prior rather than differentiating ideological nuances.
\textbf{(2) Logit Steering yields consistent and substantial improvements.}
Across both LLMs, our method boosts accuracy by 19--21 percentage points and Macro-F1 by 12--14 points.
These gains are particularly notable given that the backbone remains frozen and we train only a single linear head.
\begin{table*}[t]
\centering
\caption{\textbf{Main Results on MITweet (12 Facets).} $\Delta$ denotes improvement over the corresponding zero-shot baseline.}
\label{tab:main_results}
{\footnotesize
\begin{tabular}{lccccccc}
\hline\hline
\textbf{Model} & \textbf{Method} & \textbf{Setting} & \textbf{Accuracy} & \textbf{Macro-F1} & $\Delta$Acc & $\Delta$F1 \\
\hline
\textbf{Reference (Supervised)} & & & & & & \\
RoBERTa-base & Full FT & SFT & 59.80 & 38.50 & - & - \\
BERTweet & Full FT & SFT & 62.50 & 41.72 & - & - \\
\hline
\textbf{Qwen-2.5-7B} & & & & & & \\
{} & Zero-shot & Frozen & 44.93 & 36.55 & - & - \\
{} & Schema-Aware & Frozen & 49.25 & 32.18 & +4.32 & -4.37 \\
{} & Few-shot (5-shot) & Frozen & 48.87 & 37.87 & +3.94 & +1.32 \\
{} & \textbf{Ours} & \textbf{Frozen} & \textbf{65.88} & \textbf{48.38} & \textbf{+20.95} & \textbf{+11.83} \\
\hline
\textbf{Llama-3-8B} & & & & & & \\
{} & Zero-shot & Frozen & 46.96 & 37.43 & - & - \\
{} & Schema-Aware & Frozen & 50.82 & 33.02 & +3.86 & -4.41 \\
{} & Few-shot (5-shot) & Frozen & 54.35 & 36.98 & +7.39 & -0.45 \\
{} & \textbf{Ours} & \textbf{Frozen} & \underline{\textbf{66.83}} & \underline{\textbf{50.84}} & \textbf{+19.87} & \textbf{+13.41} \\
\hline\hline
\end{tabular}
}
\end{table*}
\subsection{Facet-Level Analysis}
Table~\ref{tab:facet_results} examines the five facets where zero-shot Llama-3 suffers the strongest degradation.
These facets exhibit severe class imbalance and high conceptual ambiguity, making them particularly sensitive to inherent model priors.
Our method consistently recovers performance, achieving gains of 20--30 points in Macro-F1.
\begin{table}[t]
\centering
\caption{\textbf{Recovery on the Most Challenging Facets (Llama-3-8B).}}
\label{tab:facet_results}
\small
\begin{tabular}{lccc}
\hline\hline
Facet & Description & Base F1 & Ours F1 ($\Delta$) \\
\hline
PeR & Personal Right & 0.0730 & \textbf{0.3730} (+0.3000) \\
MF & Military Force & 0.1400 & \textbf{0.4200} (+0.2800) \\
SS & State Structure & 0.3439 & \textbf{0.5439} (+0.2000) \\
CSR & Church–State & 0.3238 & \textbf{0.5038} (+0.1800) \\
DS & Diplomatic Strategy & 0.4843 & \textbf{0.5943} (+0.1100) \\
\hline
Avg & — & 0.2730 & \textbf{0.4870} (+0.2140) \\
\hline\hline
\end{tabular}
\end{table}

\section{Analysis}
\label{sec:analysis}

This section empirically examines the misalignment phenomena motivating our
approach and evaluates how the proposed \emph{Dual-Probe} steering mechanism
(Direction + Gravity + Uncertainty Gating) corrects these failures.
Our analysis addresses four questions:
(1) Where do ideological errors arise in frozen LLMs?
(2) Is ideological information geometrically encoded in hidden states?
(3) How do the Direction and Gravity components behave in practice?
(4) Does readout-level steering preserve the base model's linguistic integrity?

\subsection{Diagnosing Ideological Misalignment}
\label{subsec:heatmap}

\begin{figure}[t]
    \centering
    \includegraphics[width=0.9\linewidth]{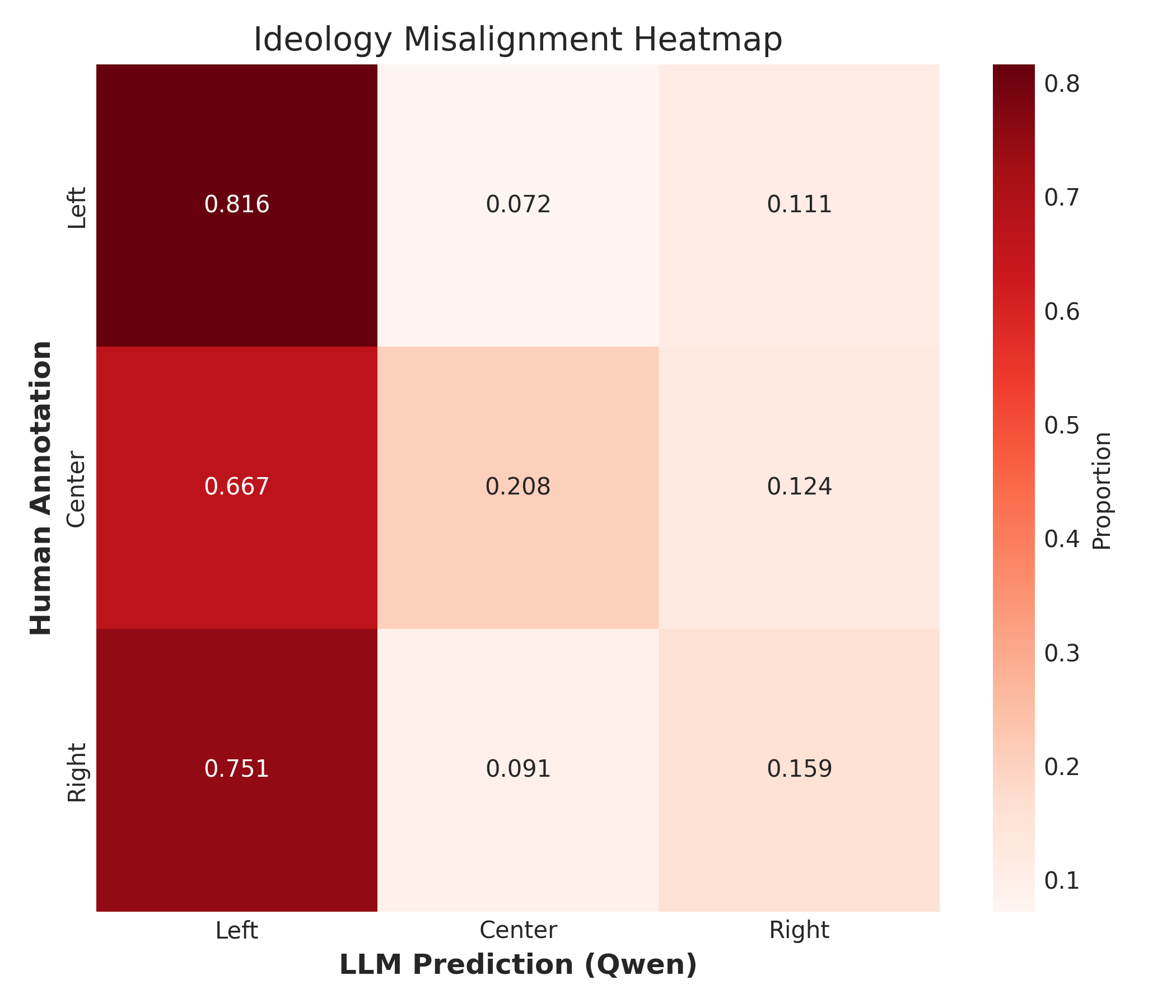}
    \caption{\textbf{Ideology Misalignment Heatmap.}  
    LLM exhibits strong prediction collapse toward the Left class}
    \label{fig:heatmap_skew}
\end{figure}

We begin by examining how a frozen LLM maps human labels to predictions.
Figure~\ref{fig:heatmap_skew} shows the row-normalized confusion matrix of
Qwen-2.5-7B under zero-shot prompting. The model exhibits a pronounced
\emph{prediction collapse}: regardless of whether the true label is Left,
Center, or Right, the model predicts \textit{Left} almost exclusively.

Importantly, this phenomenon does not imply that the model’s internal
representations are themselves uniformly left-aligned. Instead, it indicates
that the \textbf{readout layer} imposes a strong prior toward the Left class,
overriding finer ideological distinctions present in the input. This motivates a
structured readout-level calibration mechanism (Section~\ref{sec:method})
capable of correcting the biased decision surface without modifying the
underlying semantic representations.

\subsection{Geometric Structure of Ideological Representations}
\label{subsec:geometry}

\begin{figure}[t]
    \centering
    \begin{minipage}{0.48\linewidth}
        \centering
        \includegraphics[width=\linewidth]{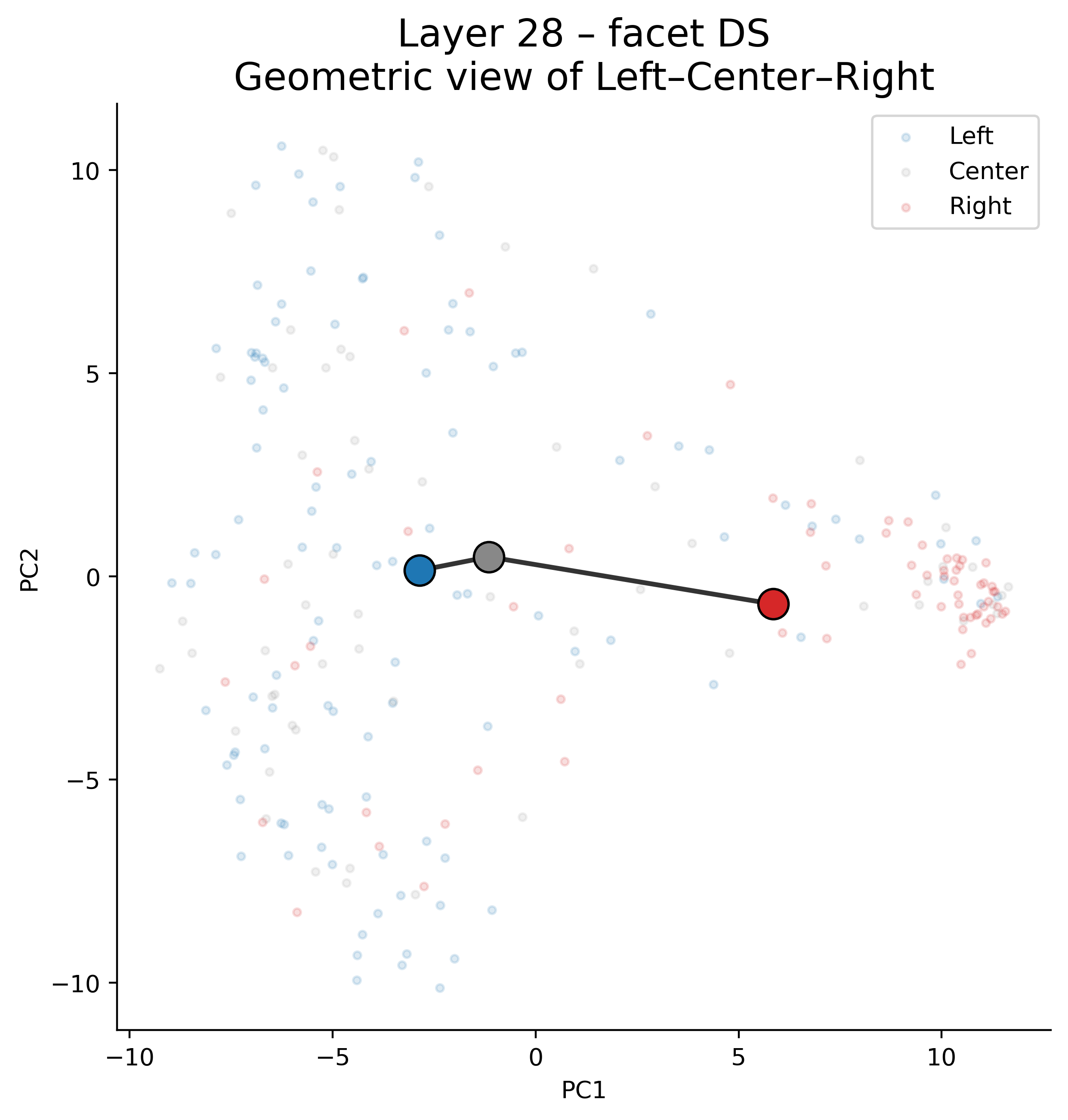}
        \centerline{\small (a) Facet: Globalization}
    \end{minipage}
    \hfill
    \begin{minipage}{0.48\linewidth}
        \centering
        \includegraphics[width=\linewidth]{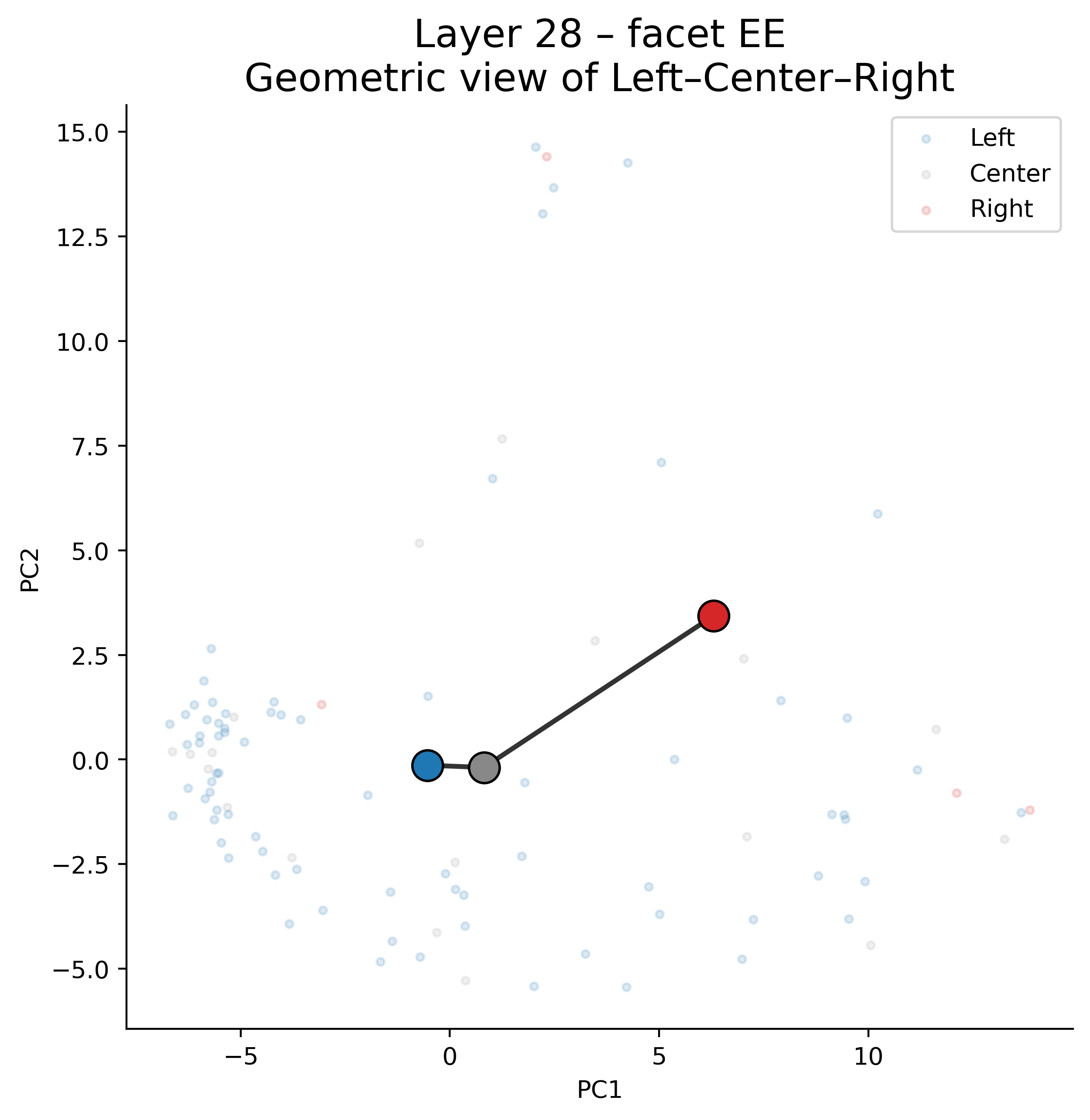}
        \centerline{\small (b) Facet: Economy}
    \end{minipage}
    \caption{\textbf{Representation Geometry.}  
    Hidden states reveal a dominant directional axis corresponding to ideological 
    variation and a concentrated uncertainty band near the center. This motivates 
    separating directional steering ($s$) from stability regularization ($g$).}
    \label{fig:manifold}
\end{figure}

To determine whether misalignment arises from the hidden representations or the
readout layer, we analyze the geometry of hidden states from Llama-3-8B and
Qwen-2.5-7B. Figure~\ref{fig:manifold} shows PCA projections for two facets
(\textit{Globalization} and \textit{Economy}) at Layer 28.

Across both models and facets, two patterns emerge:

\begin{itemize}
    \item \textbf{Ordered directional variation.}  
    The Left, Center, and Right classes form an approximately monotonic ordering 
    along a dominant direction, indicating that ideological variation is 
    \emph{geometrically encoded} in a recoverable subspace.

    \item \textbf{Residual uncertainty around the center.}  
    Center samples form a denser band orthogonal to the dominant direction,
    reflecting substantial epistemic uncertainty near the ideological midpoint.
\end{itemize}

This structure validates the design of the Dual-Probe mechanism: the Direction
probe ($s$) captures the dominant ideological axis, while the Gravity probe ($g$)
regularizes unstable regions around the center to prevent over-steering.

\subsection{Validation of Dual-Probe Decomposition Dynamics}
\label{sec:dualanalysis}

To verify the effectiveness of our \textit{Dual-Probe Decomposition} (Sec. \ref{sec:method}), we analyze the latent behaviors of the directional term $s$ (Eq. 1) and the score $g$ (Eq. 2). Specifically, we aim to confirm that the decomposition successfully disentangles the \textit{direction} of adjustment from the \textit{magnitude} of correction as hypothesized. 

We examine the internal dynamics across four distinct interaction groups:
\begin{itemize}
    \item \textbf{Group A (Aligned, $L \to L$):} Consistent samples where prediction matches ground truth.
    \item \textbf{Group B (Conflict, $R \to L$):} Samples requiring directional reversal ($s$-dominant).
    \item \textbf{Group C (Neutralization, $C \to L$):} Hallucinated bias requiring symmetric reduction ($g$-dominant).
    \item \textbf{Group D (Injection, $Side \to C$):} Stance injection into neutral predictions ($s$-dominant).
\end{itemize}

Figure \ref{fig:dual_probe_dynamics} visualizes the distributions of the optimized components.

\subsubsection{Directional Steering via Term $s$}
The distribution of the directional term $s$ (Fig. \ref{fig:dual_probe_dynamics}, Left) validates its role in capturing the signed Left-Right tendency.
\textbf{Implicit Calibration.} Both \textbf{Group A} and \textbf{Group C} show positive shifts. Since the model suffers from Left Prediction Collapse, $s$ applies a counter-force to balance the logits, consistent with the directional update $\mu s$ defined in our asymmetric calibration.
\textbf{Conflict Resolution.} \textbf{Group B} requires the largest magnitude of $s$, confirming that when the representation diametrically opposes the ground truth, the directional probe takes the primary role in shifting the logits.

\subsubsection{Orthogonality of the Score $g$}
\begin{figure}[t]
    \centering
    \includegraphics[width=\linewidth]{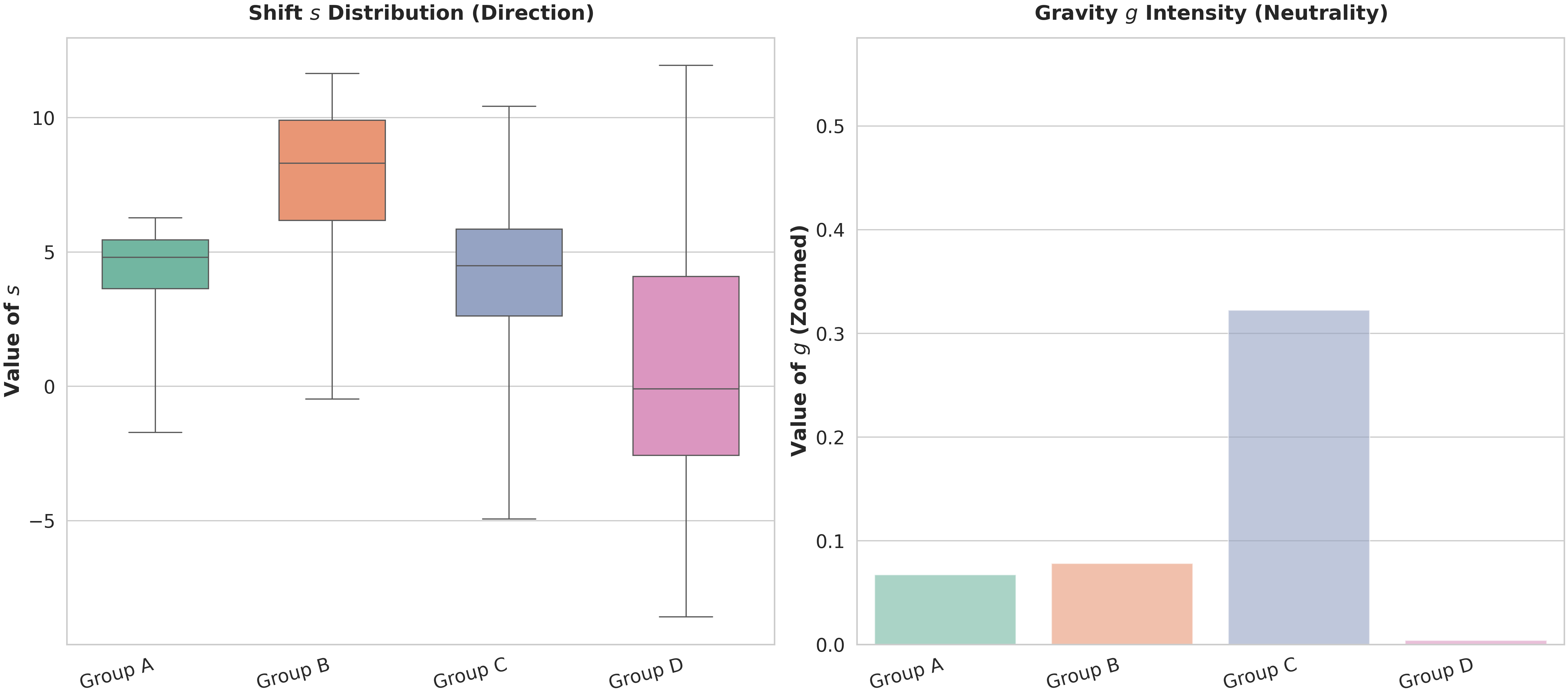}
    \caption{\textbf{Dynamics of the Dual-Probe Decomposition components.} 
    \textbf{Left:} The \textit{Directional term} $s$ shifts to counteract intrinsic bias, showing maximum activation for conflict resolution (Group B).
    \textbf{Right:} The \textit{Score} $g$ validates the disentanglement hypothesis: it activates specifically for neutralization (Group C) to apply symmetric reduction, while remaining silent during stance injection (Group D), proving it is orthogonal to directional changes.}
    \label{fig:dual_probe_dynamics}
\end{figure}

The activation of the score $g$ (Fig. \ref{fig:dual_probe_dynamics}, Right) empirically proves the disentanglement property of our architecture.
\textbf{Selective Symmetric Reduction.} As designed, $g$ spikes significantly only in \textbf{Group C} (Mean $\approx 0.32$). This confirms that Eq. (2) functions as a specific "uncertainty detector," triggering the symmetric reduction ($-\frac{1}{2}g$) only when the model hallucinates polarized features from neutral inputs.
\textbf{Independence from Direction.} Crucially, in \textbf{Group D} (Injection), $g$ remains negligible ($\approx 0.00$). Although this group undergoes a strong directional shift via $s$, it does not trigger the penalty $g$. This validates that our formulation successfully separates directional information from correction magnitude, preventing the "Score" term from interfering with legitimate stance adjustments.

\subsection{Qualitative Efficacy}
\label{subsec:qual}

Table~\ref{tab:case_examples} highlights representative corrections.  
The third example is particularly illustrative:  
although the text is clearly left-leaning, a conventional linear steering method
would risk shifting it toward the Right.  
Our dual-probe system,thus preserving the correct Left prediction.

This demonstrates that decomposing the update into two yields
more robust calibration, particularly for neutral or near-neutral content.

\begin{table}[t]
\centering
\caption{\textbf{Qualitative examples of corrected predictions.}}
\label{tab:case_examples}
\small
\begin{tabular}{p{5.0cm}cc}
\hline\hline
\textbf{Tweet Excerpt} & \textbf{Zero-shot} & \textbf{Ours} \\
\hline
``Stronger border control is the only way to restore order.'' (CV)  
& Left & \textbf{Right} \\
``Military action is necessary to defend national interests.'' (MF) 
& Left & \textbf{Right} \\
``Equal marriage rights should remain protected.'' (CV) 
& Left & \textbf{Left (Stable)} \\
\hline\hline
\end{tabular}
\end{table}

\section{Conclusion}

We introduced a lightweight and non-invasive method for aligning LLM predictions with the 
ideological preferences of human annotators. By learning a single steering direction and 
applying a simple logit-level correction, our approach improves performance on the MITweet 
benchmark without modifying model parameters or harming general capabilities. 

Our analysis shows that political facets in hidden space follow a low-dimensional structure, 
making readout-level calibration both effective and sufficient. Future work may explore 
richer corrections for facets with more complex geometry, as well as applications to other 
subjective annotation tasks.

\bibliographystyle{IEEEbib}
\bibliography{icme2026references}

\end{document}